\documentclass[journal]{IEEEtran}
\usepackage{subcaption} 
\usepackage{booktabs}
\usepackage{algorithm}
\usepackage{algorithmic}
\usepackage{amsmath}
\usepackage{amsfonts}
\usepackage{tcolorbox} 
\usepackage{newfloat}
\usepackage{listings}
\usepackage{graphicx}
\usepackage{cite}
\usepackage[pagebackref,breaklinks,colorlinks]{hyperref}
\usepackage{pifont}

\definecolor{mygray}{gray}{0.95}
\DeclareMathOperator*{\argmin}{arg\,min}
\newcommand{\etal}{\textit{et al.}}

\begin{document}

\title{Point2Quad: Generating Quad Meshes from Point Clouds via Face Prediction}

\author{Zezeng~Li$^1$,
	  Zhihui~Qi$^1$,
        Weimin~Wang$^1$,
        Ziliang~Wang$^1$,
        Junyi~Duan$^1$,
        and~Na~Lei$^1$
\thanks{This research was supported by the National Key R$\&$D Program of China under Grant No. 2021YFA1003003;  the Natural Science Foundation of China under Grant No. T2225012, No. 12494554, and No. 62306059.}
\thanks{Corresponding author: Na~Lei (Email: nalei@dlut.edu.cn).}

School of Software Technology and the DUT-RU International School of Information Science and Engineering, Dalian University of Technology, Dalian 116024, China
}

\markboth{IEEE TRANSACTIONS ON CIRCUITS AND SYSTEMS FOR VIDEO TECHNOLOGY,~Vol.~XX, No.~XX, 2024}%
{Shell \MakeLowercase{\textit{et al.}}: A Sample Article Using IEEEtran.cls for IEEE Journals}

\maketitle

\begin{abstract}
Quad meshes are essential in geometric modeling and computational mechanics. Although learning-based methods for triangle mesh demonstrate considerable advancements, quad mesh generation remains less explored due to the challenge of ensuring coplanarity, convexity, and quad-only meshes. In this paper, we present \textbf{Point2Quad}, the first learning-based method for quad-only mesh generation from point clouds. The key idea is learning to identify quad mesh with fused pointwise and facewise features. Specifically, Point2Quad begins with a k-NN-based candidate generation considering the coplanarity and squareness. Then, two encoders are followed to extract geometric and topological features that address the challenge of quad-related constraints, especially by combining in-depth quadrilaterals-specific characteristics. Subsequently, the extracted features are fused to train the classifier with a designed compound loss. The final results are derived after the refinement by a quad-specific post-processing. Extensive experiments on both clear and noise data demonstrate the effectiveness and superiority of Point2Quad, compared to baseline methods under comprehensive metrics. The code and dataset are available at \url{https://github.com/cognaclee/Point2Quad}.
\end{abstract}


\section{Introduction}
\label{sec:intro}
\IEEEPARstart{M}{any} popular acquisition techniques for 3D scans reproduce the scanned object as loose sets of 3D points. However, 
applications typically require meshes to make further use of an object. To ensure the deployment of downstream tasks, it is necessary to reconstruct the surface mesh. Existing methods are mainly focused on reconstructing triangle meshes from point clouds, but few consider the quadrangularity directly. 
{Quadrilateral meshes, however, are widely used in computational mechanics, geometric modeling, computer-aided design, computer-aided manufacturing, and simulation~\cite{lei2020quadrilateral}.
Compared to triangle meshes, quadrilateral meshes can reduce both approximation errors and the number of mesh elements~\cite{shepherd2008hexahedral,bommes2013quad}. In iso-geometric analysis, quadrilateral meshes are easier to manipulate than their triangle counterparts and have a better approximation power, benefiting from the ability to align with principal curvature directions~\cite{d2000bilinear,bommes2013quad}. Additionally,
quadrilateral meshes serve as the foundational structure for tensor-product splines, NURBS~\cite{chen2019quadrilateral,bommes2013quad}, and offer superiority in texture mapping and compression~\cite{bommes2013quad}.}

To perform quad meshing on point clouds, existing algorithms usually require a time-consuming parameterization~\cite{pietroni2011global, li2011meshless,li2011direct,jakob2015instant, schertler2017field}, KD-trees space partition~\cite{kalogerakis2009extracting,pang2014effective, bukenberger2018hierarchical}, voxels partition~\cite{guan2020voxel}, or centroidal Voronoi tessellation~\cite{levy2010p,chen2018point,zhong2019surface}. Although these traditional methods can convert point clouds to meshes to some extent, they either generate quad-dominant meshes~\cite{kalogerakis2009extracting,levy2010p,li2011direct,pang2014effective, schertler2017field,bukenberger2018hierarchical,chen2018point,zhong2019surface}, are sensitive to noise~\cite{jakob2015instant,guan2020voxel}, or cannot adapt to arbitrary topological surfaces~\cite{bukenberger2018hierarchical,guan2020voxel}. {Inspired by the powerful learning and fitting capabilities of deep learning models, we leverage a deep learning framework to generate quad meshes from point clouds, aiming for both high surface fitting accuracy and superior mesh quality.}

Unfortunately, the combinatorial nature of the quadrilateral prevents taking derivatives over the space of possible meshings of any given surface. As a result, mesh generation techniques have not been able to take full advantage of the gradient descent components of modern optimization frameworks. To bypass this non-differentiable combination problem, researchers have proposed various excellent reconstruction methods from point clouds to triangular meshes through implicit function field isosurface learning~\cite{liao2018deep, deprelle2019learning, boulch2021needrop, shen2021deep, ma2022reconstructing, williams2022neural,chen_neural_2022,boulch2022poco,jun2023shap} or mesh basic element prediction~\cite{dai_scan2mesh_2019,liu2020meshing,daroya_rein_2020,sharp_pointtrinet_2020,rakotosaona2021learning,chen2024meshanything,wang2024mergenet}. Nonetheless, they cannot be directly applied to quad meshes generation. Unlike triangle elements, which naturally satisfy coplanarity and convexity, quadrilateral requires additional constraints to meet these properties~(Eq.~\eqref{eq:coplanarity} for coplanarity,  Eq.~\eqref{eq:aspect_ratio} and Sine value of angles for convexity).

To address this challenge, we propose \textbf{Point2Quad} by reframing quadrilateralization as a local face prediction task. The key idea is to extract quadrilaterals for each point with information such as
coplanarity, convexity, and squareness. We first design a point cloud encoder to learn the geometric features from the points, and a face encoder to capture the geometric and local topological features of candidate faces. Then, we devise a prediction module to filter out geometrically and topologically reasonable quadrilaterals. Additionally, we introduce a face loss and a post-processing step to reduce holes and overlaps in the reconstructed surface, effectively enhancing the quality of the output meshes.

The main strengths of our Point2Quad can be summarized as: \textbf{quad-only mesh}: whereas other approaches may only promise quad-dominant meshes, ours is guaranteed to feature only quads. \textbf{Robustness to noise}: our method can filter out noise when reconstructing the meshes. \textbf{High reconstruction accuracy}: the reconstructed meshes have high accuracy.
In summary, our main contributions are:
\begin{itemize}
        \item For the distinct challenge of quad meshing, we introduce the first learning-based point-to-quad method that generates high-quality quad-only meshes via face prediction.
        \item We present an effective method to encode geometric and topological features for quad extraction by considering position and coplanarity, convexity, squareness of faces.
        \item To address the class imbalance and the 
        potential for holes and overlaps, we introduce a compound loss and effective post-processing.
	\item We build a medium-size quad mesh dataset. Qualitative and quantitative results under comprehensive metrics show the superiority of our method to SOTA baselines.	
\end{itemize}

\section{Related Works}
There is extensive work for mesh generation, thorough surveys can be found in reference~\cite{ ho1988finite,bommes2013quad,lei2023s}. Following, we only briefly review the most related works.

\subsection{Learing-based point-to-triangle approaches}  
A variety of recent works focus on reconstructing triangle meshes from point clouds in a learning way. Unlike image-to-mesh generation~\cite{wang2018pixel2mesh,wen2019pixel2mesh++,li2021pcgan,wang2022progressive,chen2022transformer,huang2023simultaneously,li2023high,yao2024staf}, the primary challenge in point cloud-to-mesh generation is determining the vertex connections, which is essentially a combinatorial problem. Additionally, point clouds are often sparse, noisy, or incomplete, further increasing the difficulty of mesh generation from point clouds.

To address these challenges, researchers have proposed various approaches  \cite{li_incomplete_2008,badki2020meshlet,daroya_rein_2020,yifan2021iso,boulch2021needrop,ma_neural-pull_2021,ma2022reconstructing,williams2022neural,lin2022multiview,wang2024mergenet}. Li \etal~\cite{li_incomplete_2008} are the first to consider generating meshes from incomplete point clouds using neural networks. Badki \etal~\cite{badki2020meshlet} handles sparse or noisy point clouds by learning local geometric priors while ensuring global geometric consistency. Alexandre and Ma \etal~\cite{boulch2021needrop, ma_neural-pull_2021,ma2022reconstructing} construct highly accurate meshes from sparse point clouds by pulling query 3D locations to their closest points on the surface. Wang \etal~\cite{wang2024mergenet} introduced a method for reconstructing meshes from sparse point clouds by predicting edge connections. This approach extracts features from candidate edges and predicts their distances to the target surface, effectively filtering out the true edges. Manifold mesh generation~\cite{rakotosaona2021differentiable,chen_neural_2021,Rakotosaona_2021_CVPR,peng2021shape,chen_neural_2022,maruani2023voromesh} is often achieved by combining Delaunay triangulation or the Marching Cubes (MC)~\cite{lorensen1987marching}. For example, Chen \etal~\cite{chen_neural_2021} introduced the first MC-based approach capable of recovering sharp geometric features.  Rakotosaona \etal~\cite{Rakotosaona_2021_CVPR} leveraged 2D Delaunay triangulations to construct a manifold mesh. Lin \etal~\cite{lin2022multiview} take advantage of a differentiable Poisson Solver (SAP)~\cite{peng2021shape} to reconstruct meshes. Nissim \etal~\cite{maruani2023voromesh} proposed a Voronoi-based method to reconstruct closed and non-self-intersecting meshes from noisy point clouds.

From a technical standpoint, these methods primarily address the combination problem in mesh generation through two main approaches. The first is implicit function field isosurface learning~\cite{liao2018deep, deprelle2019learning, boulch2021needrop, shen2021deep, ma2022reconstructing, williams2022neural,chen_neural_2022,boulch2022poco,lin2022multiview,jun2023shap,maruani2023voromesh}, and the second is mesh basic element prediction~\cite{dai_scan2mesh_2019,liu2020meshing,daroya_rein_2020,sharp_pointtrinet_2020,rakotosaona2021learning,chen2024meshanything,wang2024mergenet}. Our proposed Point2Quad method is a derivative of the basic element prediction approach. We treat quad mesh generation as a face selection problem, bypassing the non-differentiable nature of combinatorial tasks by classifying candidate surfaces.

\subsection{Learing-based quad mesh generation} 
To take advantage of the powerful representation capability of deep neural networks, some researchers have also tried to combine deep learning with quad mesh generation and have obtained good solutions \cite{smirnov2020learning,dielen2021learning,chen2022mgnet,deng2022sketch2pq,pan2023reinforcement,zhou2023quadrilateral,tong2023srl,li2024craftsman}.

By fitting a deformable parametric template composed of coons patches, Smirnov et al.~\cite{smirnov2020learning} decomposite target surfaces into patches. Then, quadrilateral meshes are obtained by simply subdividing. Another interesting work is LDFQ \cite{dielen2021learning}, which learns cross fields from a triangular mesh. LDFQ can infer frame fields that resemble the alignment of quads and produce high-quality quad meshes, but it requires convoluted data pre-processing and post-processing. MGNet~\cite{chen2022mgnet}, an unsupervised structured mesh generation method, uses boundary curves as input and employs a neural network to learn the parameterization map of the target surface, generating meshes with the desired number of faces. However, it needs to retrain the network for each target surface. Sketch2PQ~\cite{deng2022sketch2pq} addresses quad mesh generation for open surfaces by predicting the corresponding spline surface and conjugate direction field from sketches. CraftsMan~\cite{li2024craftsman} takes a different approach by leveraging a diffusion model to generate triangular meshes from text prompts or reference images, later converting them to quadrilateral meshes using an off-the-shelf remeshing tool.

By formulating the mesh generation as a Markov decision process problem, Pan et al.~\cite{pan2023reinforcement} provided a reinforcement learning-based method to automatically learn the policy of actions for mesh generation for any given plane boundary. Tong \etal~\cite{tong2023srl} extend this method by incorporating singularity configurations into the reward function, resulting in more structured meshes. However, both methods are primarily designed for planar mesh generation. To generate block-structured quadrilateral mesh for 2D geometric models, Zhou et al.~\cite{zhou2023quadrilateral} proposed to identify the singular structure contained in the frame field. After that, the segmentation streamline is constructed between singularities, and the domain is decomposed into several small regions that are finally discretized with quadrilateral mesh elements. After obtaining the singular structure output by the neural network, this method requires meticulous post-processing to extract streamlines, especially for domains with complex topology and geometry.

Despite these advancements, none of these methods directly address quadrilateral mesh generation from point clouds, leaving a gap in the current research landscape.

\subsection{Point-to-quad mesh approaches}
Most existing quad mesh generation methods require triangular meshes as input, following a remeshing approach from triangular meshes to quad meshes. Recently, with the increasing popularity of point cloud scanning devices, research on directly reconstructing quad meshes from point clouds has gained attention, resulting in several excellent works. 

These methods are generally divided into two categories: parameterization-based and space partition-based approaches. Parameterization-based methods often require time-consuming processes~\cite{pietroni2011global, li2011meshless,li2011direct,jakob2015instant, schertler2017field} to compute parameterized coordinates or cross fields. Space partition-based methods typically employ KD-tree partitioning~\cite{kalogerakis2009extracting,pang2014effective, bukenberger2018hierarchical} or voxel partitioning~\cite{guan2020voxel} to divide the target surface into regular quadrilateral blocks, followed by quad mesh generation via integer programming. Centroidal Voronoi tessellation (CVT) is another space-partitioning technique successfully applied to isotropic and anisotropic remeshing. CVT-based methods~\cite{levy2010p,chen2018point,zhong2019surface} iteratively compute Voronoi diagrams during energy optimization. Unfortunately, these methods either generate quad-dominant meshes~\cite{kalogerakis2009extracting,levy2010p,li2011direct,pang2014effective, schertler2017field,bukenberger2018hierarchical,chen2018point,zhong2019surface}, are sensitive to noise~\cite{jakob2015instant,guan2020voxel}, or cannot adapt to arbitrary genus surfaces~\cite{bukenberger2018hierarchical,guan2020voxel}.

\begin{figure*}[htbp!]
\vspace{-3mm}
\centering
\setlength{\abovecaptionskip}{0.cm}	   
\setlength{\belowdisplayskip}{2pt} 	
\includegraphics[width=\linewidth]{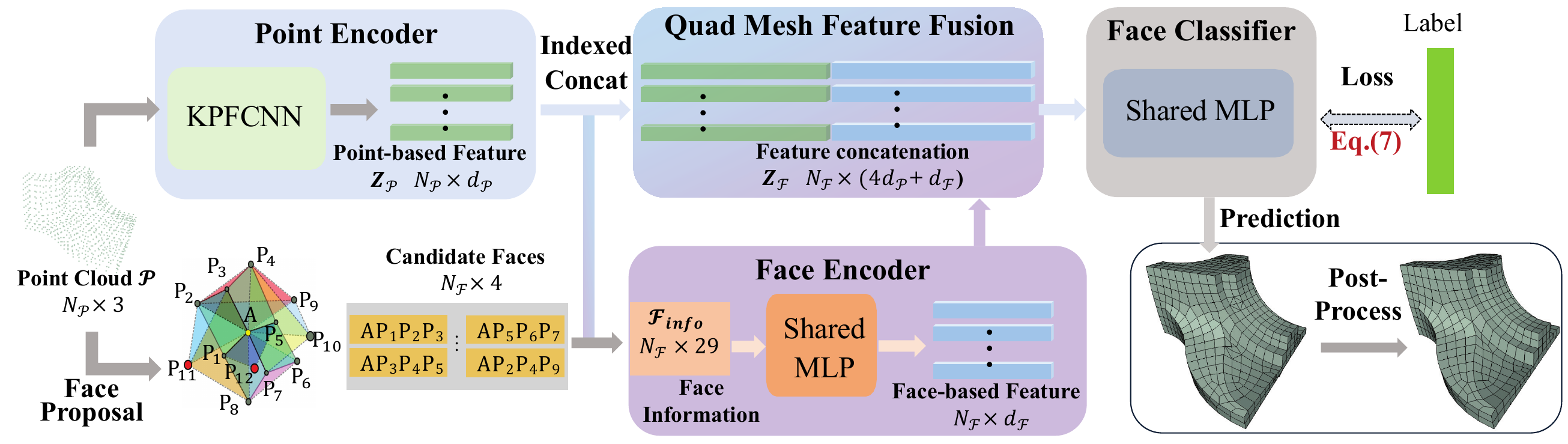}
\vspace{-3mm}
\caption{Full pipeline of our \textbf{Point2Quad}: given a point cloud as input, we first propose a set of candidate quadrilateral. During training~(dashed arrow), the network is trained to classify the candidates with the supervision of face labels. During inference, the predicted label is used to filter out quadrilaterals, which are then merged into the output mesh with a post-process.}
\label{fig:framework}
\end{figure*}

Compared to existing methods, our approach tackles the problem from a completely different angle—a local connectivity prediction. By selecting candidate quadrilaterals, we can generate quad-only meshes with high {surface fitting accuracy} and mesh quality from noisy point clouds.

\section{Methodology}
Given a point cloud $\boldsymbol{\mathcal{P}} =\left\{\boldsymbol{p}_i\right\}_{i=1}^{N_{\mathcal{P}}}$ with $N_{\mathcal{P}}$ points, our objective is to reconstruct a quadrilateral mesh, which consists of a set of vertex and a set of quadrilateral faces. Notably, most points in $\boldsymbol{\mathcal{P}}$ serve as vertices for the reconstructed quadrilateral mesh. A small portion of points are noise points, generated by adding noise in the normal direction to the sampling points of target surface~(red points $\boldsymbol{p}_{11}$ and $\boldsymbol{p}_{12}$ in the bottom-left corner of Fig.~\ref{fig:framework}).
Faces are predicted as connections between each point and its neighbors. We first construct a $k$-nearest neighbor ($k$-NN) graph for each point in $\boldsymbol{\mathcal{P}}$ based on the Euclidean distance, and then construct candidate quadrilateral faces centered on each point. Finally, a neural network is utilized to extract suitable quadrilaterals from these overlapping candidate quadrilaterals. The pipeline of our method is shown in Fig.~\ref{fig:framework}.

\subsection{Problem formulation}
\label{sec:formulation}
Given a reference quadrilateral mesh $\mathcal{S}_\mathcal{R}$ and a point cloud $\boldsymbol{\mathcal{P}}$ sampled on $\mathcal{S}_\mathcal{R}$ as input, Point2Quad aims to generate a quad mesh $S_\mathcal{P}$ whose vertices come from the point cloud $\boldsymbol{\mathcal{P}}$. As shown in Fig.~\ref{fig:framework}, Point2Quad first proposes $N_{\mathcal{F}}$ candidate faces $\boldsymbol{\mathcal{F}}_{\mathcal{P}}$ and then uses a subset of candidate faces to form the mesh $\mathcal{S}_\mathcal{P}$. Given the true face label $\boldsymbol{\mathcal{Y}}$ of the reference mesh,
our model aims to fit a function of the points and their neighbors to the face label, as shown in Eq.~(\ref{eq:formulation}).
\begin{equation}
	\label{eq:formulation}
	g: (\boldsymbol{\mathcal{P}},\mathcal{\boldsymbol{\mathcal{F}}}_{\mathcal{P}})\rightarrow \boldsymbol{\mathcal{Y}}\,.
\end{equation}
Where $\boldsymbol{\mathcal{Y}}$ is a $N_{\mathcal{F}}$ dimension vector of ground truth face labels, $\boldsymbol{\mathcal{P}}$ is a coordinate matrix with the shape of $N_{\mathcal{P}}\times 3$.

\subsection{Candidate face generation}
To get these candidate quad faces, we first construct a $k$-nearest neighbor ($k$-NN) graph for each of the points in $\mathcal{\boldsymbol{P}}$ based on the Euclidean distance, and then each point can propose a set of candidate quadrilateral face with each three of its $k$-NN neighbors. By constructing candidate quadrilaterals in this way, the reconstructed mesh can also adapt to different point cloud resolutions. Among all the candidates, there exists a specific subset that precisely forms a quadrilateral mesh. In this subset, each vertex forms a no-overlapping quad loop with its neighbors. We need to filter out false candidates and use the remaining faces to obtain the mesh $S_\mathcal{P}$. If each non-boundary vertex constitutes a quadrilateral ring with its neighbors, a watertight mesh can be obtained.

\begin{figure}[htb]
\vspace{-3mm}
\centering
\includegraphics[width=0.6\linewidth]{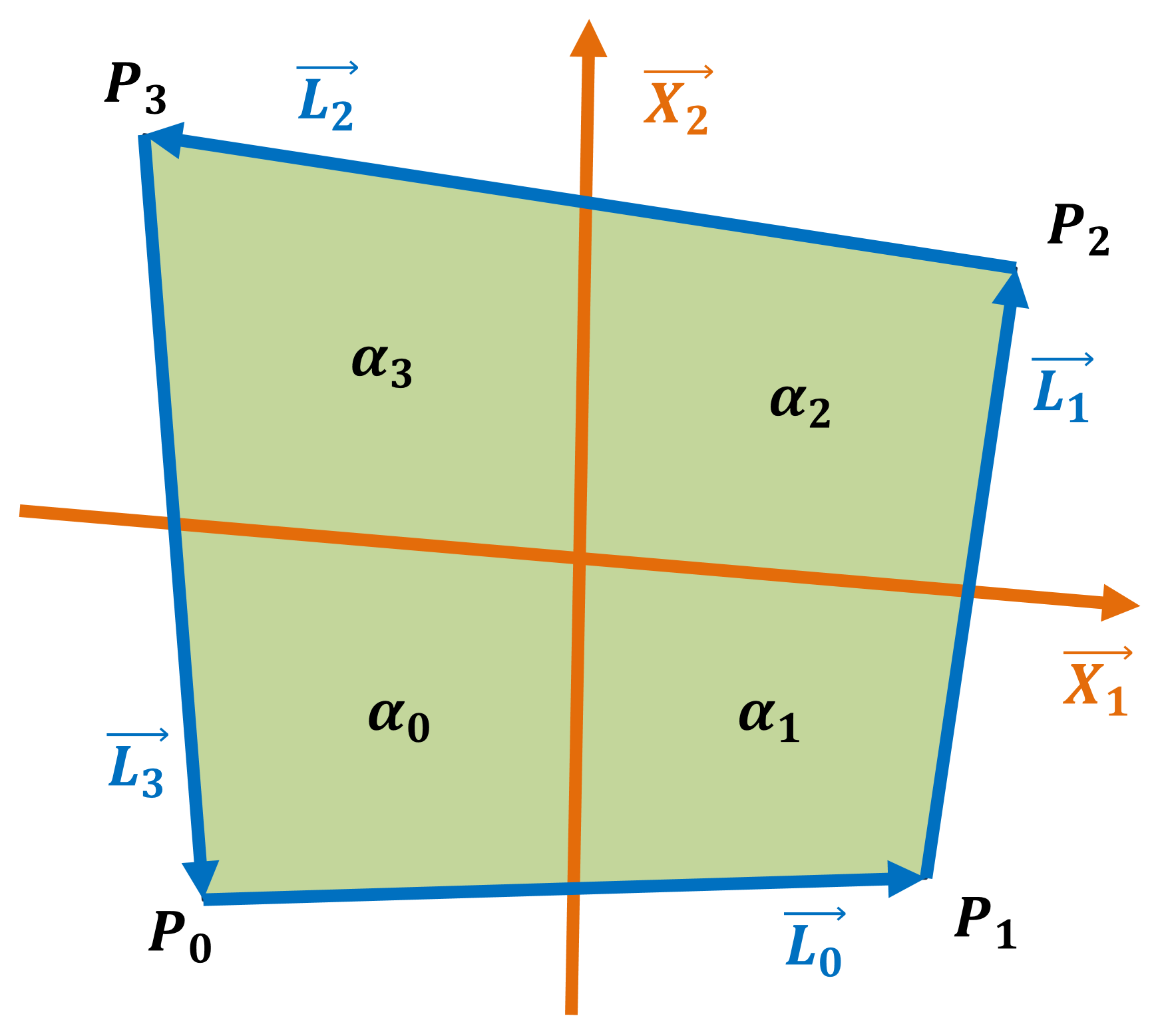}
\vspace{-4mm}
\caption{Illustration of notions in quadrilaterals.}
\label{fig:scaledJacobian}
\end{figure}
To cover the whole surface of $\mathcal{S}_\mathcal{R}$, we first use a sufficiently large $k$ to construct the neighbors for each point, then select the quadrilaterals being reasonable in geometry as candidate faces. Specifically, we filter them with coplanarity, aspect ratio, and orthogonality.  We calculate the dot product of the normals of each pair of vertexes in the face, and then filter the faces based on threshold value. If the dot product of the normals on a quadrilateral face is small or negative, it indicates that the coplanarity of the face is poor. Given the candidate surface $f$ as shown in Fig~\ref{fig:scaledJacobian}, the normal of each vertex can be calculated by the cross product of its two adjacent edges:
\begin{equation}
\boldsymbol{n}(\boldsymbol{p}_i) = \Vec{L_i} \times\Vec{L}_{i \bmod 4}.
\label{eq:coplanarity}
\end{equation}
In addition, we get the reciprocal of the aspect ratio as :
 \begin{equation}
r(f) = \frac{\mathop{\min}_{i\in\{0,1,2,3\}}\Vert \Vec{L}_{i} \Vert}{\mathop{\max}_{i\in\{0,1,2,3\}} \Vert \Vec{L}_{i} \Vert},
\label{eq:aspect_ratio}
\end{equation}
whose value ranges from 0 to 1. The reciprocal of the aspect ratio serves as an indicator of facial stretching, where values that converge towards 0 signify a heightened level of distortion. Furthermore, we evaluate the orthogonality of the face by computing the sine values at the four corners of the face. A sine value that approximates 0 underscores a decline in the face's orthogonality. We impose thresholds on these metrics to ensure that unreasonable candidates are excluded.

By employing these filtering strategies, we identify candidate faces that exhibit acceptable properties of convexity, coplanarity, and squareness, thereby eliminating the majority of invalid candidates from the potential vertex combinations. While this approach significantly reduces the number of candidate faces (from $\mathcal{O}(N_{\mathcal{P}}^4)$), selecting appropriate quadrilateral faces to form a mesh remains challenging. Two primary issues persist: non-manifold conditions arising from overlapping faces and non-watertight results due to missing faces. The following section explores how to select an appropriate subset from overlapping quadrilateral candidates to construct a mesh that maximizes both watertightness and manifoldness.

\subsection{Pointwise and facewise features extraction}
To facilitate the subsequent selection of candidate quadrilaterals, it is essential to first encode the input point cloud. Point clouds are rich in geometric features that significantly influence the layout of the quadrilateral mesh. A natural approach is to utilize a deep neural network built on the point cloud for geometric feature extraction. Given the effectiveness of KPConv~\cite{thomas2019kpconv} in capturing local geometry, we employ KPConv as the backbone network for point encoder $\boldsymbol{Enc}_{\mathcal{P}}$. Then, as shown in Fig.~\ref{fig:framework}, we index the feature corresponding to each vertex's identity number in the target quadrilateral and concatenate them to get the geometric feature of the quadrilateral, formally expressed as
\begin{equation}
\begin{split}
\boldsymbol{Z}_{\mathcal{P}}&= \boldsymbol{Enc}_{\mathcal{P}}(\boldsymbol{\mathcal{P}})\,,\\
\boldsymbol{Z}_{\mathcal{F}}^{\mathcal{P}}&= \boldsymbol{Z}_{\mathcal{P}}(\boldsymbol{\mathcal{F}}_{\mathcal{P}})\,.
\label{eq:geometric}
\end{split}
\end{equation}
Where $\boldsymbol{Z}_{\mathcal{P}}$ is the pointwise features with size $N_{\mathcal{P}}\times d_\mathcal{P}$. $\boldsymbol{Z}_\mathcal{F}^{\mathcal{P}}$ represents the geometric features of candidate faces with size $N_{\mathcal{F}}\times 4d_\mathcal{P}$.

However, geometry alone is not sufficient to determine a quad mesh. Candidate quadrilaterals carry local topological connectivity information, which can provide crucial guidance for the subsequent selection of quadrilaterals. To this end, we utilize additional network modules $\boldsymbol{Enc}_{\mathcal{F}}$ to encode facewise information, formally expressed as:
\begin{equation}
\label{eq:topological}
\boldsymbol{Z}_{\mathcal{F}}^{\mathcal{F}}= \boldsymbol{Enc}_{\mathcal{F}}(\boldsymbol{\mathcal{F}}_{info})\,,
\end{equation}
where $\boldsymbol{Z}_\mathcal{F}^{\mathcal{F}}$ denotes the facewise features with size $N_{\mathcal{F}}\times d_\mathcal{F}$. $\boldsymbol{\mathcal{F}}_{info}$ is a $N_{\mathcal{F}}\times 29$ matrix that carries the local topological connectivity of candidate faces and geometry of vertexes. Each row of $\boldsymbol{\mathcal{F}}_{info}$ is composed of the coordinates~(4$\times$3) of four vertices, a scaled Jacobian value~(1$\times$1)~\cite{knupp2000achieving}, the sine values~(4$\times$1) of four interior angles, and the normal vectors~(4$\times$3) of the four vertices within a candidate face in $\boldsymbol{\mathcal{F}}_{\mathcal{P}}$. Among these,  the sine values of the four interior angles provide insights into the convexity and squareness of candidate faces, while the normal vectors of vertices reflect its coplanarity. Scaled Jacobian $J_s$ measures the coplanarity and squareness of a face, calculated as:
\begin{equation}
\label{eq:Jacobian}
J_s = \mathop{\min}_{i\in\{0,1,2,3\}}\frac{\alpha_i} {\Vert \Vec{L_i} \Vert \Vert \Vec{L}_{i \bmod 4} \Vert} \,,
\end{equation}
where $\alpha$ represents the area of one of the four parts of a quadrilateral and $\Vec{L}$ is an edge vector, as shown in Fig.~\ref{fig:scaledJacobian}. 

Based on the adjacency topology information and geometric information such as convexity, coplanarity, and squareness provided by $\boldsymbol{\mathcal{F}}_{info}$, the encoder can extract sufficient information to help the Face Classifier to choose those candidate surface with greater convexity, coplanarity, and squareness, thus complete the subsequent face classification task accurately.

\subsection{Training and prediction}
After obtaining the geometric and local features mentioned above, the next task is to select the correct quadruple to form a high-quality quad mesh. To achieve this, we first concatenate the two features to obtain the overall feature $\boldsymbol{Z}_\mathcal{F} = [\boldsymbol{Z}_\mathcal{F}^{\mathcal{P}};\boldsymbol{Z}_\mathcal{F}^{\mathcal{F}}]\in\mathbb{R}^{N_{\mathcal{F}}\times(4d_{\mathcal{P}}+d_{\mathcal{F}})}$. Given the excellent integration of geometric and topological information in this feature representation, we only need a simple Face Classifier~(see Fig.~\ref{fig:framework}), such as a shared weight multiple layer perceptron (MLP), to screen for the appropriate quadrilaterals. The convolutional operation and the shared weight MLP are designed to inspire the learning of generalizable local priors across different parts and shapes. For details on the network and hyperparameter settings, Please refer to Section~\ref{sec:details}.

In training, we have ground truth mesh and follow the steps in Section~\ref{sec:formulation} to obtain candidate faces and their corresponding labels, which serves as dense supervision for candidate face prediction. Specifically, candidates are divided into 2 classes in our experiments. Candidates who are on the ground truth surface are in class 1, otherwise, are in class 0. The network $g_{\Theta}$ is thus trained to predict a 2-class label for each candidate in an end-to-end fashion with a compound loss:
\begin{equation}
\label{eq:compound}
\argmin_{\Theta}\, \,  \mathcal{L}_{C} (g_{\Theta} (\boldsymbol{\mathcal{P}},\boldsymbol{\mathcal{F}}_{\mathcal{P}}), \boldsymbol{\mathcal{Y}})+\mathcal{L}_{F} (g_{\Theta} (\boldsymbol{\mathcal{P}},\boldsymbol{\mathcal{F}}_{\mathcal{P}}), \boldsymbol{\mathcal{Y}}).
\end{equation}

First, a weighted cross-entropy loss $\mathcal{L}_{C}$ between the per-face prediction  $\hat{y}=g_{\Theta} (\boldsymbol{\mathcal{P}},f_{\mathcal{P}})$ and the ground truth label $y$ is adopted to guide the optimization of model parameters,
\begin{equation}
\label{eq:crossEntropy}
\mathcal{L}_{C} = -\sum_{i=1}^M \left[w y_i \log(\hat{y}_i) + (1 - y_i) \log(1 - \hat{y}_i)\right].
\end{equation}
The weight $w$ here can be adjusted based on the ratio of 0 and 1 labels in the training dataset. We found through preliminary experiments that a slightly bigger value of $w$ compared to the ratio of 0 and 1 labels would be better. 

Further, to reduce holes and overlaps in meshes, face loss 
\begin{equation}
\label{eq:faceloss}
\mathcal{L}_{F} = -\sum_{i=1}^M \left[y_i \exp(\hat{y}_i^0) + (1 - y_i) \exp(\hat{y}_i^1)\right],
\end{equation}
is introduced. Where $\hat{y}_i^0$ and $\hat{y}_i^1$ represent the probability outputs on class 0 and class 1, respectively.

At the inference stage, we use the predicted labels to filter out the incorrect candidates and then sort the remaining candidates according to their labels. We finally merge them into the output mesh through a post-process algorithm.

\subsection{Post process}
\begin{figure}[H]
\centering
\includegraphics[width=1.0\linewidth]{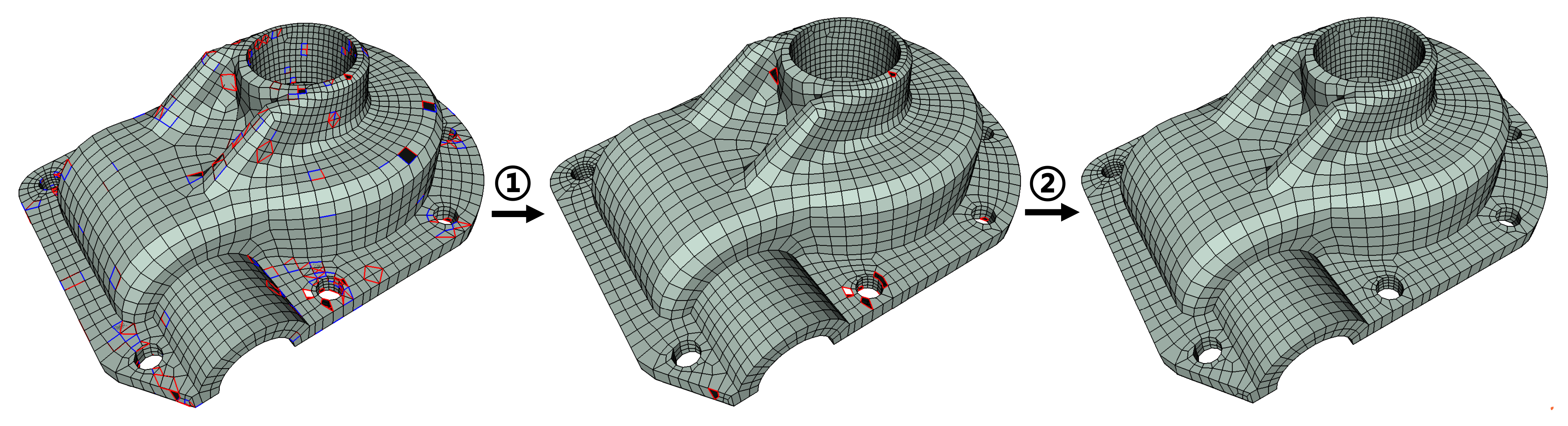}
\caption{post-process: \ding{172} remove non-manifold parts with scores in Eq.~\eqref{eq:FaceScore}; \ding{173} fill holes automatically. Non-manifold and boundary edges are highlighted in \textcolor{blue}{blue} and \textcolor{red}{red}.}
\label{fig:post_process_pipeline}
\end{figure}

Despite the impressive result of the network, some faces overlap, which makes the mesh nonmanifold. The post-process is to handle the problem, output ameliorated and neat quad meshes. We first remove the overlapped elements that cause some small holes in the surface, then fill holes with a greedy method, the workflow shown in Fig.~\ref{fig:post_process_pipeline}.

In the post process, we first remove the overlapped elements that cause some small holes in the surface, then fill the hole with a greedy algorithm. For each quadrilateral face, we score it based on the degrees of its four edges, and then delete the faces with high scores in descending order based on their scores. For each quadrilateral face $f$, the score
\begin{equation}
\label{eq:FaceScore}
E_s(f) = (E_b(f)+1)*(10*E_n(f)+1),
\end{equation}
where $E_b(f)$ and $E_n(f)$ denote the number of boundary edges and non-manifold edges in $f$. 
Obviously, $E_s(f) \geq 1$, and $E_s(f) = 1$ only if $E_b(f)$ and $E_n(f)$ are both 0, which means the face is regular. Similarly, the face containing only one boundary edge corresponds to $E_s(f) = 2$, which should not be deleted to avoid endless deletions. 

\begin{figure}[H]
\centering
\includegraphics[width=0.4\linewidth]{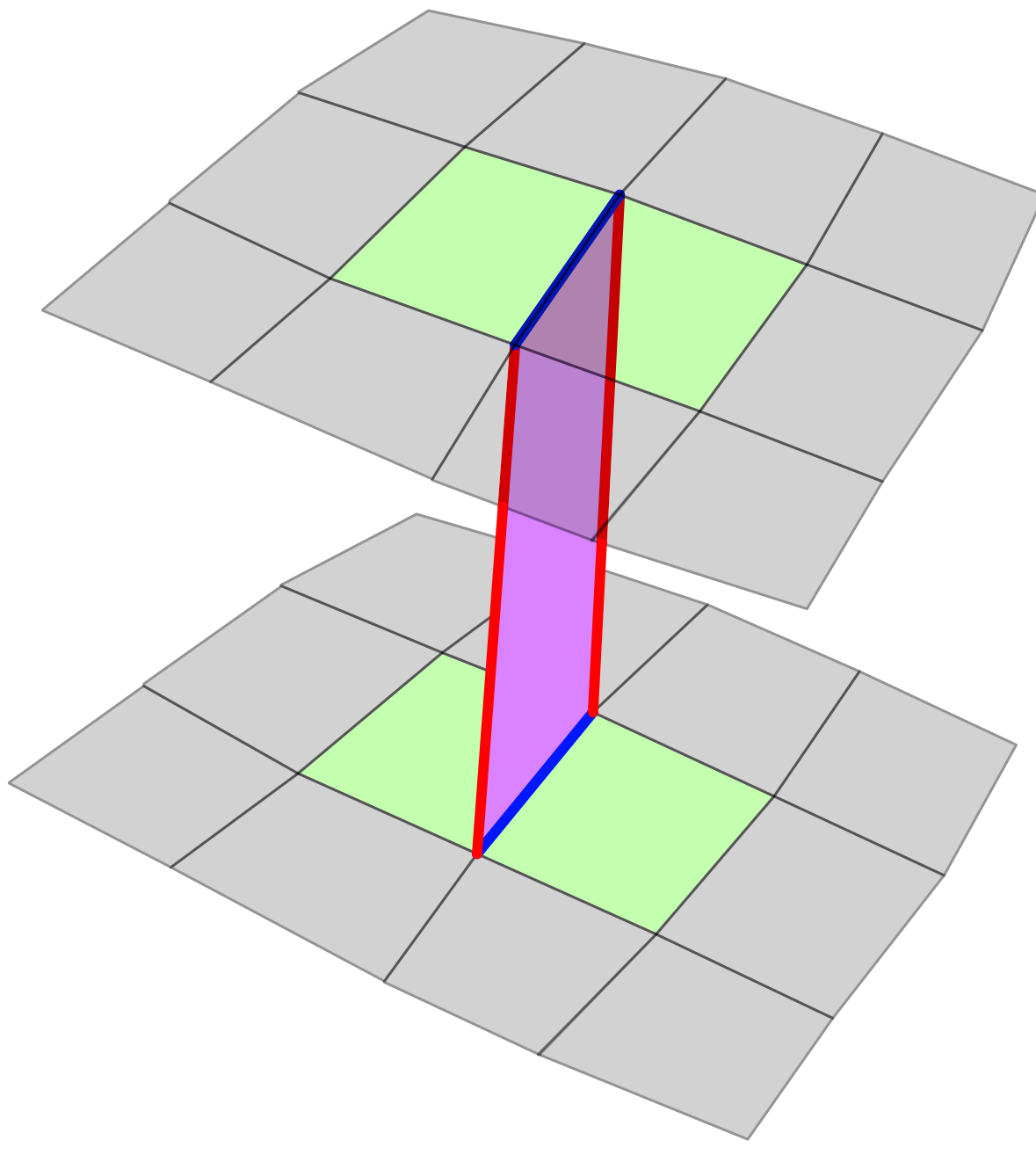}
\caption{Scoring example: The \textcolor{blue}{blue} and \textcolor{red}{red} edges are non-manifold and boundary edges respectively.}
\label{fig:manifold}
\vspace{-3mm}
\end{figure}

We remove faces in descending order until only faces with scores of 2 or 1 remain. During the process, the scores of the affected elements are simultaneously refreshed. The proposed score $E_s$ in Eq.~\eqref{eq:FaceScore} prioritizes the removal of faces with high non-manifold degrees, while in non-manifold cases, faces with more boundaries are removed earlier. As illustrated in the scoring example in Fig.~\ref{fig:manifold}, the purple face, which has a higher score, will be deleted before the green faces. Upon deleting the purple face, the green face becomes a regular face, thereby reducing unnecessary deletions. Experimental results have shown that this method can reduce the non-manifold edges while rarely causing extra unreasonable holes.

\begin{figure}[htb]
\vspace{-3mm}
\centering
\includegraphics[width=0.6\linewidth]{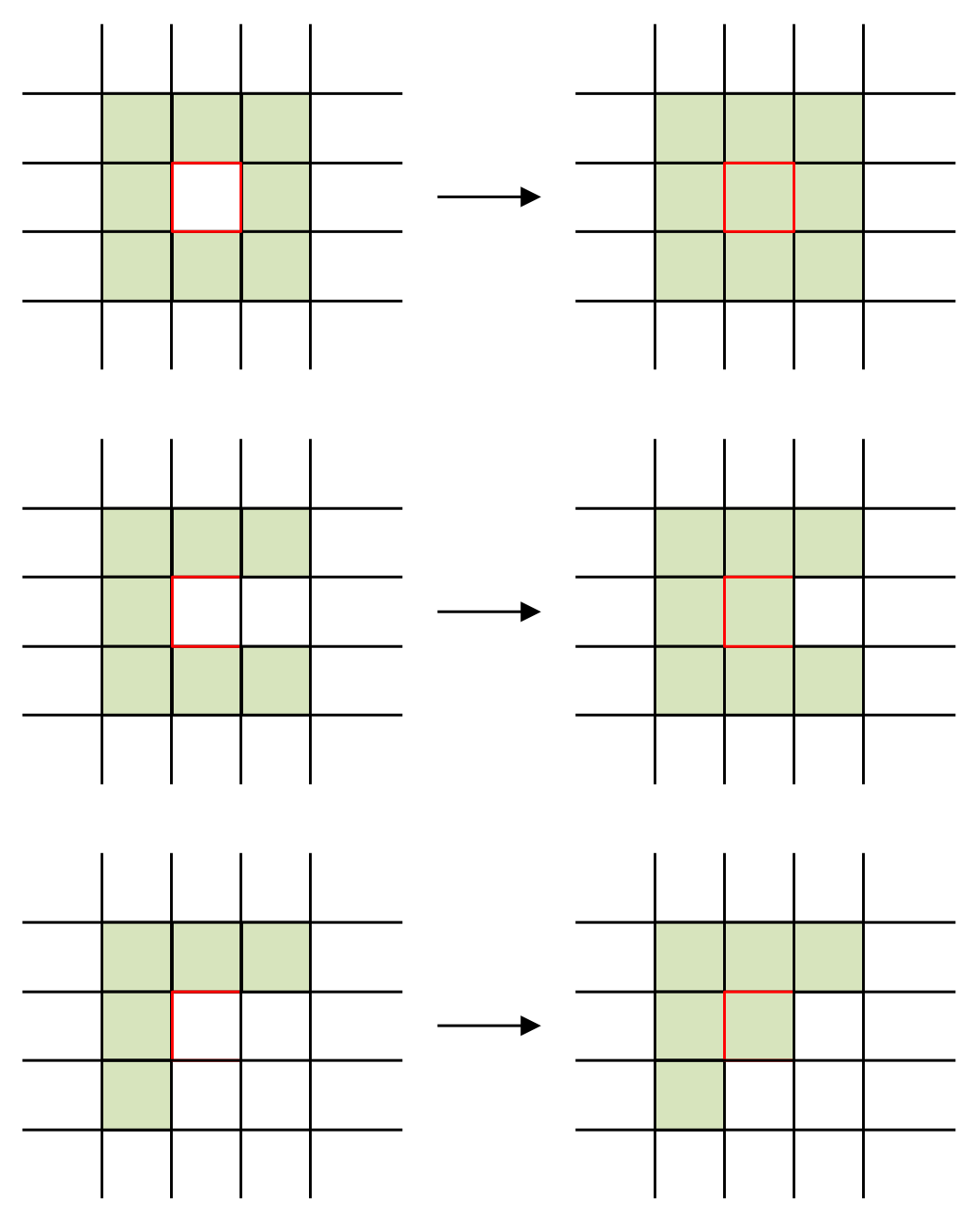}
\caption{Fillhole strategy in greedy algorithm}
\label{fig:fillhole}
\end{figure}

\begin{figure*}[htb!]
\centering
\subcaptionbox*{Point cloud
\qquad\qquad\qquad Ground truth \qquad\qquad\qquad Instant Meshes
 \qquad\qquad\qquad\quad IER \qquad\qquad\qquad \textbf{Ours}}
{\includegraphics[width=\textwidth,height=20.cm]{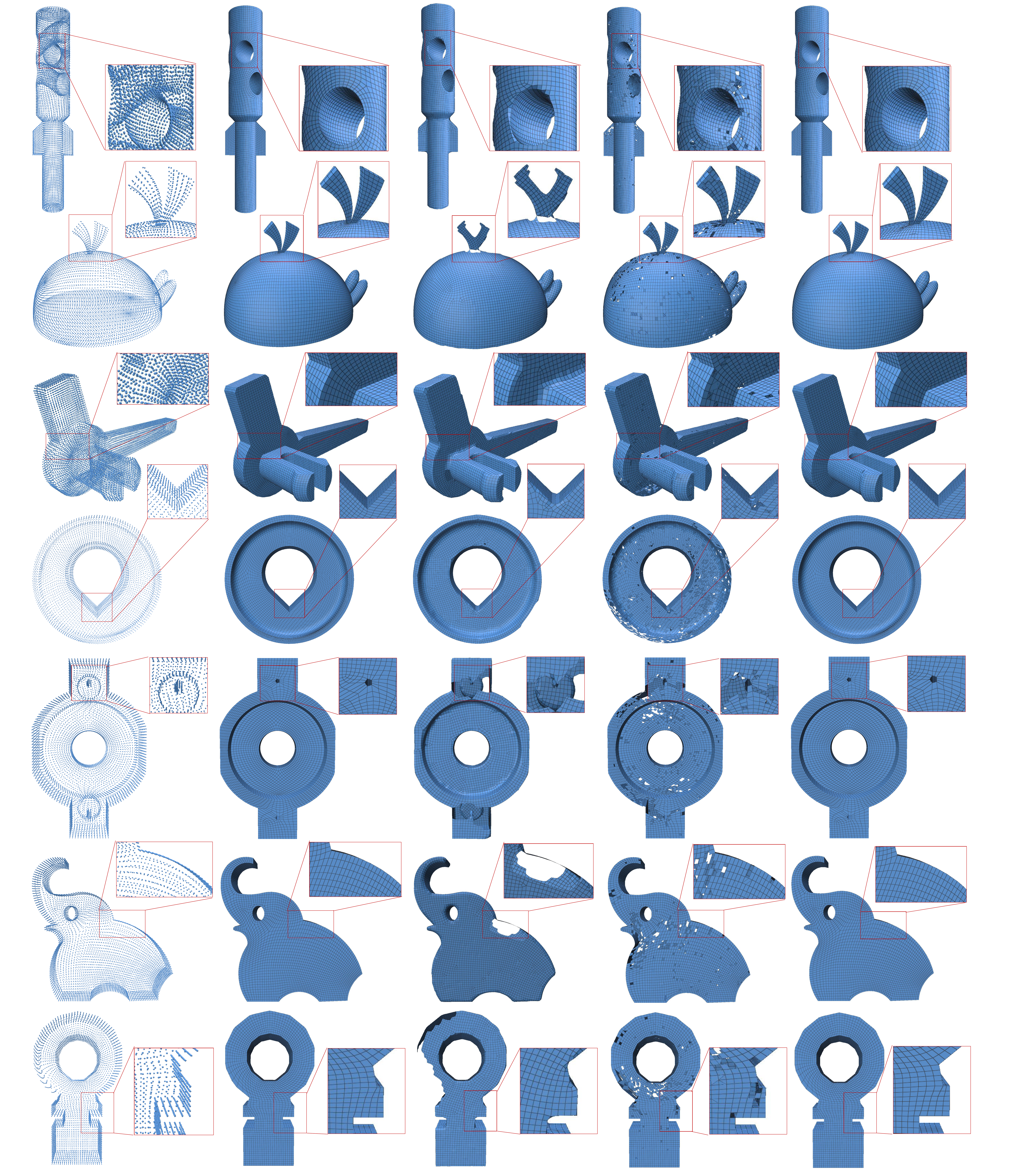}}
\caption{Visualization results and close-up views of quad meshes generated from noisy point clouds.}
\label{fig:qualitative}
\vspace{-3mm}
\end{figure*}

Then we fill holes through a greedy algorithm. Specifically, we attempt to identify 3 patterns by tracing the boundary edges along the boundary holes, as shown in Fig.~\ref{fig:fillhole}. In the simplest case, if 4 consecutive boundary edges form exactly one hole, then we only need to create a new face here to fill it. Secondly, when we encounter three consecutive boundary edges that are accompanied by two corners close to right angles, we identify this configuration as a concave hole segment. To address this, we fill the concave hole by connecting the source point of the first edge with the target point of the last edge, effectively bridging the gap and restoring the boundary's continuity. Additionally, if two consecutive edges form an angle that is close to a right angle, we identify this as a boundary angle. In such cases, we create a new point by bilinear interpolation of the edge vectors. This new point is then connected to both edges, thereby enhancing the structural integrity and precision of the boundary. This greedy method works similarly to the advancing front method~\cite{schoberl1997netgen}, gradually filling the hole from the border inward.

\section{Experiments}
In this section, we present comprehensive experiments to evaluate the efficacy of our proposed method. We first introduce the dataset, comparison methods, and metrics, and then present the quantitative and qualitative comparison in Section~\ref{sec:evaluation}. Finally, we conduct ablation studies in Section~\ref{sec:ablation}.

\vspace{-2mm}
\subsection{Dataset} \label{sec:dataset}
We construct our dateset \textit{Point\&Quad} by a series of high-quality mesh as ground truth. The corresponding point clouds are constructed by the vertices of meshes with about 10\% noise points. The label is based on candidate faces, for each vertice, we get the candidate faces, then we label the candidate faces 0 or 1  whether it exists in the mesh. Considering the memory limitations of GPUs, we randomly obtained 1,641 models with no more than 10,000 vertexes from the quadrilateral version of Thing10K dataset~\cite{pietroni2021reliable}. The quad meshes in the dataset are watertight and manifold, covering 0-genus, high-genus, organic, and CAD models with sharp features, etc. 
We randomly split the dataset into 1,312 models (80\%) for training and 312 models for testing. All quantitative results reported in this section represent average scores across the entire test set.

\begin{figure}[htb]
\centering
\includegraphics[width=1.0\linewidth]{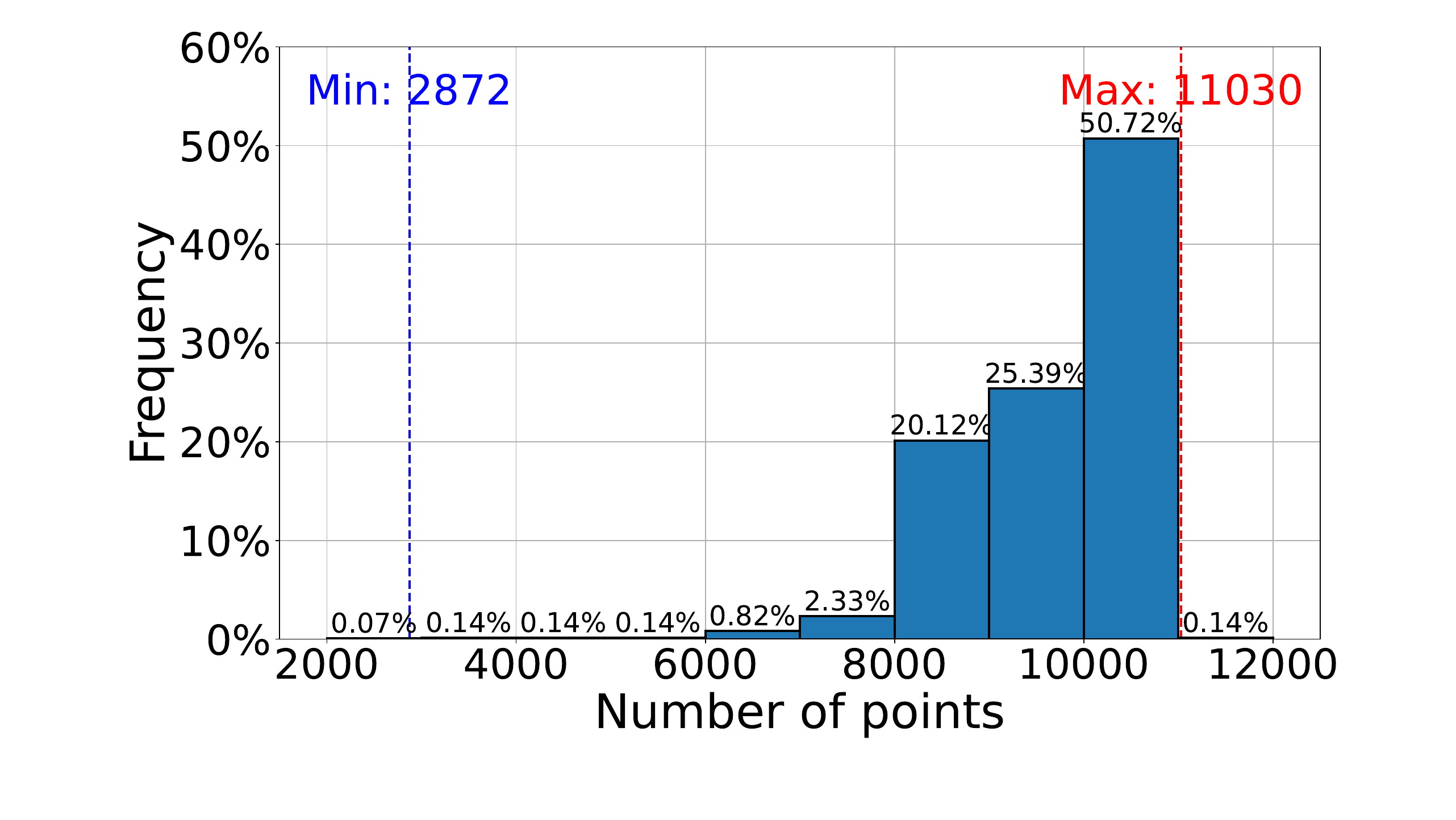}
\vspace{-6mm}
\caption{Distribution of the number of points for each 3D object in our dataset. The minimum number of points is 2872, and the maximum is 11030.}
\label{fig:distribution}
\end{figure}

To generate noise into point clouds, we first randomly select $N$ faces based on the size of the face area, assuming  $N$ noise points are needed. Next, we obtain 
$N$ points on the object's surface through random interpolation. Each of these points is then randomly perturbed by multiplying the face's normal by a random scalar and adding it to the sampled point. After adding noise,  the distribution of the number of points in the point clouds of our dataset is shown in Fig.~\ref{fig:distribution}.
\vspace{-3mm}

To generate candidate quadrilateral faces for subsequent processing, we first calculate the Euclidean distance between every pair of points in the quadrilateral mesh and sort these distances in ascending order. For each point, $k$ closest points are selected as neighbor points. From these neighbors, we randomly choose three points and arrange them in counterclockwise order with the original point. Geometric filters based on elongation, orthogonality, and coplanarity are then applied to eliminate unreasonable faces. Specifically, we compute the min-max edge ratio, the sine of angles, and the dot product of normals to evaluate these geometric properties, applying thresholds of 0.25, 0.3, and 0.5, respectively, to filter the faces. The faces that pass this filtering process are considered candidate faces, resulting in 12 candidate faces for each point.

\subsection{Evaluation Metric} \label{sec:Metric}
We quantitatively evaluate the output mesh in terms of both shape fit and mesh quality. For quad mesh quality, we use the generic scaled Jacobian~\cite{knupp2000achieving}, max-min edge ratio~\cite{knupp2006verdict}, angle distortion~\cite{knupp2006verdict} as a quantitative metric. For {surface fitting accuracy}, we use the Chamfer distance~\cite{hanocka2020point2mesh} to measure the proximity between the input points and 10K points sampled on a generated mesh guided by the area of the quadrilateral, and use precision and recall to measure the ability to filter true quad faces from all candidate faces. In addition, to evaluate the manifoldness and watertightness, we adopt the metrics proposed by Nicholas \etal~\cite{sharp2020pointtrinet}, 
\begin{equation}
\begin{aligned}
    &manifoldness = \frac{\#E_1+\#E_2}{\#E_{all}},\\\
    &watertightness = \frac{\#E_2}{\#E_{all}}.
\end{aligned}
\end{equation}
Where $\#E_1$ and $\#E_2$ denote the number of edges that only appear in a quadrilateral and two quadrilateral respectively. $\#E_{all}$ denotes the total number of edges.

\begin{table}[htbp]
\caption{Metrics used in our experiment.}
\footnotesize
\renewcommand{\arraystretch}{1.0}
\renewcommand{\tabcolsep}{0.9mm}
\centering
\begin{tabular}{c|c|c|c}
\hline
Metric&Range&Expectation&Measured content\\ 
\hline
Scaled Jacobian &$[-1, 1]$ & 1& shape regularity\\
Max-min edge ratio & $[1, +\infty]$& 1 & elongation\\
Angle distortion  &$[0^{\circ}, 90^{\circ}]$& $0^{\circ}$& orthogonality\\
Watertightness  &$[0, 1]$& 1& degree of watertight\\
Manifoldness &$[0, 1]$& 1& degree of manifold\\
Chamfer distance & $[0, +\infty]$& 0 & fidelity\\
Precision/Recall& $[0, 1]$& 1 & accuracy rate\\
\hline
\end{tabular}
\label{tab:Metric}
\end{table}

The metrics we used for quantitative analysis are listed in the Tab.~\ref{tab:Metric}. The table delineates the range, expected value (values of a unit quadrilateral), and measured content for each metric utilized. Specifically, the \textit{Scaled Jacobian} is calculated by Eq.~\eqref{eq:Jacobian}. The \textit{Max-min edge ratio} refers to the ratio of the longest edge to the shortest. Both metrics are evaluated per face, with the overall score for the mesh model derived by averaging the results across all faces. 
Additionally, the \textit{Angle distortion}~(AD) is determined by calculating the mean square error between all interior $N_{\theta}$ angles and $90^{\circ}$,
\begin{equation}\label{eq:Angledistortion}
AD = \frac{1}{N_{\theta}}\sum_{\theta_i}(\theta_i-90^{\circ})^2.
\end{equation}
\textit{Chamfer distance}~(CD) is computed based on the point set $\boldsymbol{\mathcal{P}}$ from the target quad mesh and $\tilde{\boldsymbol{\mathcal{P}}}$ from the generated  mesh,
\begin{equation}
CD =\frac{1}{\#\tilde{\boldsymbol{\mathcal{P}}}}\sum_{\tilde{\boldsymbol{p}} \in \tilde{\boldsymbol{\mathcal{P}}} } \min_{\boldsymbol{p} \in \boldsymbol{\mathcal{P}}} \|\tilde{\boldsymbol{p}}-\boldsymbol{p}\|_2 + \frac{1}{\#\boldsymbol{\mathcal{P}}}\sum_{\boldsymbol{p} \in \boldsymbol{\mathcal{P}} } \min_{\tilde{\boldsymbol{p}} \in \tilde{\boldsymbol{\mathcal{P}}}} \|\boldsymbol{p}-\tilde{\boldsymbol{p}}\|_2 .
\end{equation}

\vspace{-3mm}
\subsection{Implementation details and Baselines} \label{sec:details}
\noindent\textbf{Implementation detail} We instantiated the point encoder in Fig.~\ref{fig:framework} with KPFCNN~\cite{thomas2019kpconv} and its default parameter settings in the official source code~\footnote{\url{ https://github.com/HuguesTHOMAS/KPConv}}. The feature dimension of the KPFCNN output is  $d_{\mathcal{P}}=64$. For the face encoder, we implemented it with five layers of \textit{Conv1d+Norm1d+LeakyReLU} and one residual module. The input feature dimensions for the five \textit{Conv1d} layers are 29, 128, 128, 256, and 512, respectively. The residual module follows the second \textit{Conv1d+Norm1d+LeakyReLU} layer, with an input feature dimension of 128. The output feature dimension of the Face Encoder is $d_{\mathcal{F}}=256$. The classifier comprises five layers of \textit{Conv1d+Norm1d+ReLU}, with input feature dimensions of 512, 512, 256, 128, and 64 for the respective \textit{Conv1d} layers.

In our $k$-NN setting, $k=12$. That is, each point finds its 12 nearest neighbors through Euclidean distance. Then, based on these neighboring points, 12 candidate faces are proposed with each point as the center, which means $N_{\mathcal{F}}=12N_{\mathcal{P}}$.

We train our model for 500 epochs and use a momentum gradient descent optimizer with a momentum of 0.99, an initial learning rate of $10^{-3}$. Our learning rate is scheduled to decrease exponentially, and we choose the exponential decay to ensure it is divided by 10 every 150 epochs.

\noindent\textbf{Baselines.} We compare our method against Instant-Meshes~\cite{jakob2015instant} and IER~\cite{liu2020meshing}. Note that Instant-Meshes cannot directly generate quadrilateral meshes from pure point clouds. So we first use the ball-pivoting algorithm~\cite{bernardini1999ball} to generate triangular meshes, and then use Instant Meshes to remesh the triangular meshes into quadrilateral meshes. For IER, we use their open-source code and retrain it on our datasets.

\begin{table}[ht!]
\caption{{Qualitative comparison with $10\%$ noise. $\uparrow^a_{b}$ and $\downarrow^a_{b}$ denote that larger and smaller values are better, respectively. Here, $a$ and $b$ represent the upper and lower bounds.}}
\footnotesize
\renewcommand{\arraystretch}{1.2}
\renewcommand{\tabcolsep}{1.2mm}
\centering
\begin{tabular}{l|c|c|c}
\hline
Model&Instant-Meshes & IER& \textbf{Ours}\\ 
\hline
Scaled Jacobian $\uparrow^1_{-1}$ &0.95835  & 0.87237&\textbf{0.98613} \\
Max-min edge ratio $\downarrow^{\infty}_{1}$ & 217.919& 59.0826 &\textbf{1.19228} \\
Angle distortion $\downarrow^{90}_{0}$ &11.6315 & 18.5870&\textbf{7.47548}  \\
Watertightness $\uparrow^1_{0}$ & 0.94294 & 0.60823&\textbf{0.99998} \\
Manifoldness $\uparrow^1_{0}$&\textbf{1.0} &0.99979&0.99999  \\
Chamfer distance $\downarrow^{\infty}_{0}$ & 0.04514& 0.02092&\textbf{0.02064} \\
Precision/Recall $\uparrow^1_{0}$ & -/-	& 0.86/0.94& \textbf{0.94/0.99}\\
{Time(s) $\downarrow^{\infty}_{0}$} & {2.64}	& {\textbf{1.58}}& {2.31}\\
\hline
\end{tabular}
\label{tab:Qualitative}
\end{table}

\begin{table}[ht!]
\caption{Qualitative comparison without noisy points.}
\footnotesize
\renewcommand{\arraystretch}{1.2}
\renewcommand{\tabcolsep}{1.2mm}
\centering
\begin{tabular}{l|c|c|c}
\hline
Model&Instant-Meshes & IER& \textbf{Ours}\\ 
\hline
Scaled Jacobian $\uparrow^1_{-1}$ & 0.96659 &0.89368 &\textbf{0.99053} \\
Max-min edge ratio $\downarrow^{\infty}_{1}$ &137.925 & 1.32215 &\textbf{1.18949} \\
Angle distortion $\downarrow^{90}_{0}$ &9.68795 &10.4283 &\textbf{8.04664}  \\
Watertightness $\uparrow^1_{0}$ & 0.94536 &0.66628 &\textbf{0.99995} \\
Manifoldness $\uparrow^1_{0}$&\textbf{1.0} &\textbf{1.0}&\textbf{1.0}  \\
Chamfer distance $\downarrow^{\infty}_{0}$ &0.02961 &0.02079&\textbf{0.02057} \\
Precision/Recall $\uparrow^1_{0}$ & -/-	& 0.94/0.98& \textbf{0.95/0.996}\\
\hline
\end{tabular}
\label{tab:noiseRobustness}
\end{table}

\vspace{-2mm}
\subsection{Quantitative and qualitative evaluation}\label{sec:evaluation}
\noindent\textbf{Quantitative Results.} 
We report the results in Tab.~\ref{tab:Qualitative}. As shown, Point2Quad achieved the best scores on all metrics except for manifoldness, and we also obtained a score very close to 1 on manifoldness. This indicates that Point2Quad not only generates high-quality quad meshes but also exhibits superior potential for surface fitting accuracy. Instant-Meshes~\cite{jakob2015instant} performs perfectly under manifoldness and is slightly worse than our results under the Jacobi metric. However, it performs far worse in terms of max-min edge ratio and Chamfer distance. Based on our observations, there are two main reasons for this discrepancy. Firstly, Instant-Meshes aims to create square meshes and avoid overlapping, which can result in more holes. Secondly, smoothing the cross-field inevitably leads to surface distortion and loss of details, thereby reducing {surface fitting accuracy}. IER demonstrates high surface fitting accuracy but suffers from overlap and small holes, resulting in low Chamfer distance. Additionally, its scores on other mesh quality metrics are far inferior to those of the two compared methods.

To test the robustness to noise, we evaluated the performance of each method with~(Tab.~\ref{tab:Qualitative}) and without noise~(Tab.~\ref{tab:noiseRobustness}). It can be found that our method decreases slightly, whereas Instant-Meshes and IER significantly decline. 
Instant-Meshes lacks the ability to identify noise points and indiscriminately uses all points to fit the surface. As a result, the generated meshes deviate from the true surface, exhibiting significant Chamfer distances. Similarly, IER fails to selectively process points when extracting faces, inevitably producing highly irregular quadrilaterals. This leads to a higher Max-min edge ratio and angle distortion. {We also measured the time cost of each method, as shown in the last row of Tab.~\ref{tab:Qualitative}. The results indicate that while IER holds a slight advantage, our method achieves a comprehensive improvement in mesh quality with only a minimal time cost of 0.7 seconds.}

Additionally, we compared all methods across all subintervals based on the size of point clouds in Tab.~\ref{tab:pointRobustness}. Our method consistently achieves stable performance across different point resolutions, indicating that it exhibits better robustness to variations of point cloud scale.

\begin{table}[ht!]
\caption{Robustness to the number of points. IM represents Instant-Meshes. $\geq$10K denotes the number of points more than 10K. Each block from top to bottom represents the score under scale Jacobian, max-min edge ratio, angle distortion, watertightness, manifoldness, and Chamfer distance.}

\footnotesize
\begin{center}
\renewcommand{\tabcolsep}{1.5mm}
\begin{tabular}{c|c|c|c|c|c}
\hline
Method  & 2K-4K & 4K-6K & 6K-8K & 8K-10K  & $\geq$10K  \\
\hline
IM  & 0.92556 & 0.93887&0.96154 & 0.97697&0.95436 \\
IER & 0.76279 &0.89425 &0.89037  &0.85010 & 0.90254\\
\textbf{Ours}  &\textbf{0.98149} & \textbf{0.99184}& \textbf{0.98452} &\textbf{0.98499} &\textbf{0.98755} \\
\hline
IM  & \textbf{1.29703} &\textbf{1.09711} &195.793 &230.379 &217.457 \\
IER & 3.44613 & 2.37171&475.825  &59.0899 &60.1348 \\
\textbf{Ours}  & 3.23562 &2.05991 &\textbf{1.51028}  &\textbf{1.25833} &\textbf{1.13609} \\
\hline
IM  & 12.5334 &7.13513 &\textbf{8.03783} &10.94724 &12.11786 \\
IER & 45.5038 & 33.9379& 27.0812 & 21.9391&17.6993 \\
\textbf{Ours}  & \textbf{8.40754} & \textbf{5.66304}&8.07632  &\textbf{7.85801} &\textbf{7.10932} \\
\hline
IM  &0.89332  & 0.83125& 0.90340&0.94231 &0.94305 \\
IER & 0.59668 &0.58236 &0.60991  &0.60317 &0.61889 \\
\textbf{Ours}  & \textbf{0.99812} & \textbf{0.99615}& \textbf{0.99952} & \textbf{0.99999}&\textbf{0.99998} \\
\hline
IM  & \textbf{1} & \textbf{1}& \textbf{1}&\textbf{1} &\textbf{1} \\
IER &0.99954  & 0.99963& 0.99960 &0.99978 &0.99985 \\
\textbf{Ours}  & \textbf{1} & \textbf{1}& \textbf{1} &\textbf{1} &0.99999 \\
\hline
IM  & 0.22016 &0.05329 &0.03739 & 0.04210&0.04099 \\
IER & 0.02902 & 0.01819& 0.01806 &0.02148 &0.02075 \\
\textbf{Ours} & \textbf{0.02266} &\textbf{0.01737} & \textbf{0.01736} & \textbf{0.02082}&\textbf{0.02049} \\
\hline
\end{tabular}
\label{tab:pointRobustness}
\end{center}
\end{table}

\begin{figure*}[htb!]
\centering
\subcaptionbox*{Point cloud
\qquad\qquad\qquad Ground truth \quad\qquad\qquad Instant Meshes
 \qquad\qquad\qquad IER \qquad\qquad\qquad\qquad\qquad  \textbf{Ours}\qquad}
{\includegraphics[width=\textwidth,height=20.cm]{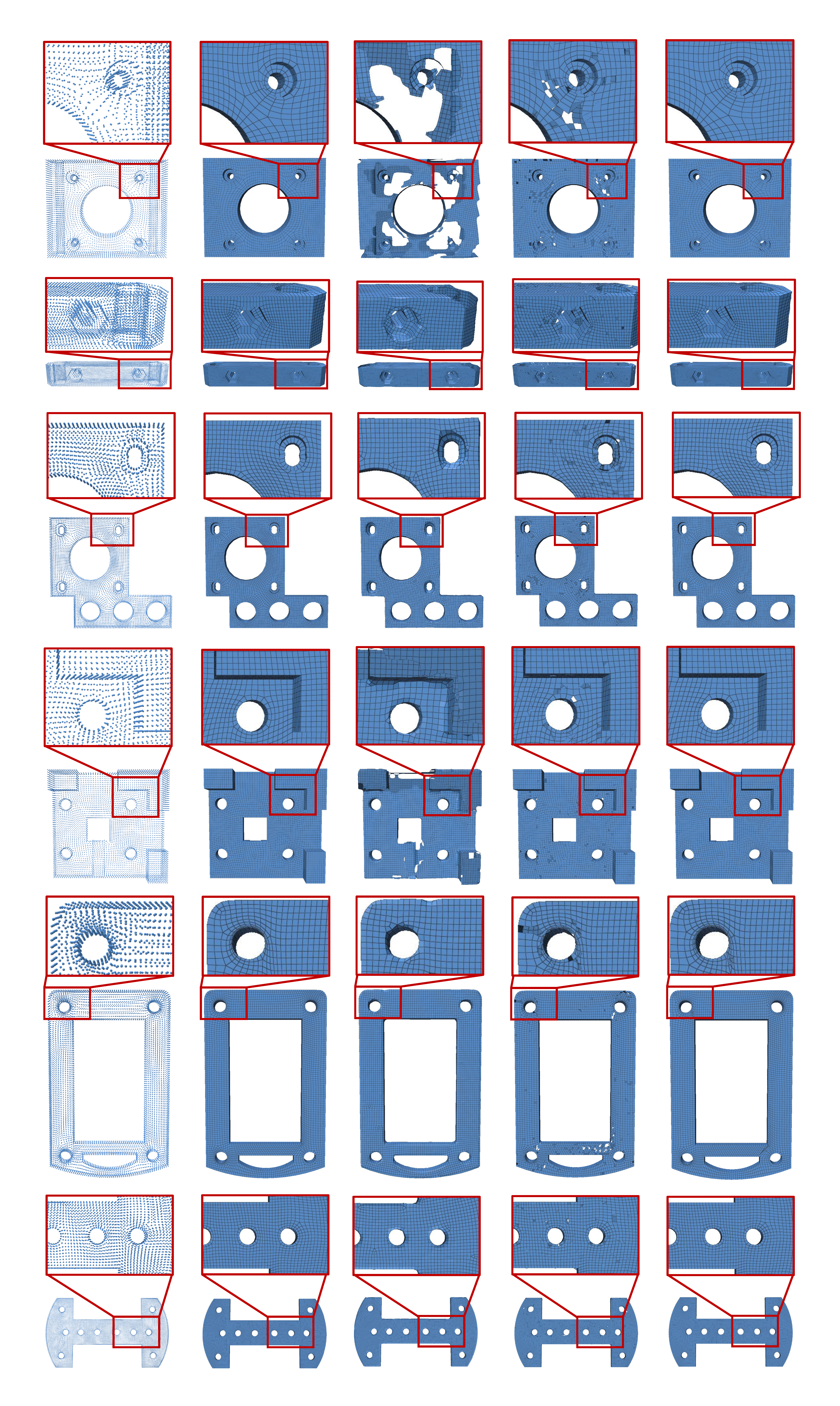}}
\caption{Visualization results and close-up views of generated quad meshes in high-genus scenarios.}
\label{fig:qualitative_supp}
\vspace{-4mm}
\end{figure*}

\noindent\textbf{Qualitative Results.} The reconstructed quad meshes by different methods are visualized in Fig.~\ref{fig:qualitative}. The results show that Instant-Meshes exhibits large missing pieces in areas where the point cloud is sparse or the curvature is high. IER's results feature many non-manifold elements, overlaps, and small holes. Our method, while maintaining an accurate fit to the target surface, achieves high mesh quality. By incorporating face-based topological information, our approach adapts well to regions with high curvature, as demonstrated in the results from the second to the fourth rows.

Furthermore, we visualized the generated results in high-genus situations.  As shown in Fig.~\ref{fig:qualitative_supp}, close-up views with significant differences are displayed above each model.  The results show that Instant Meshes~\cite{jakob2015instant} tends to produce non-watertight quad meshes with poor adherence to the object’s surface. IER~\cite{liu2020meshing} generally has more small holes and non-manifold issues. In contrast, the proposed method generates meshes with better surface adherence and excels in feature preservation and watertightness. This indicates that even in cases of high genus, the proposed method can effectively reconstruct quadrilateral meshes from point clouds.

\begin{figure}[ht!]
\centering
\subcaptionbox*{w/o FE\qquad\qquad w/o FL \qquad\qquad w/o PP \qquad\qquad Full}
{\includegraphics[width=1.0\linewidth]{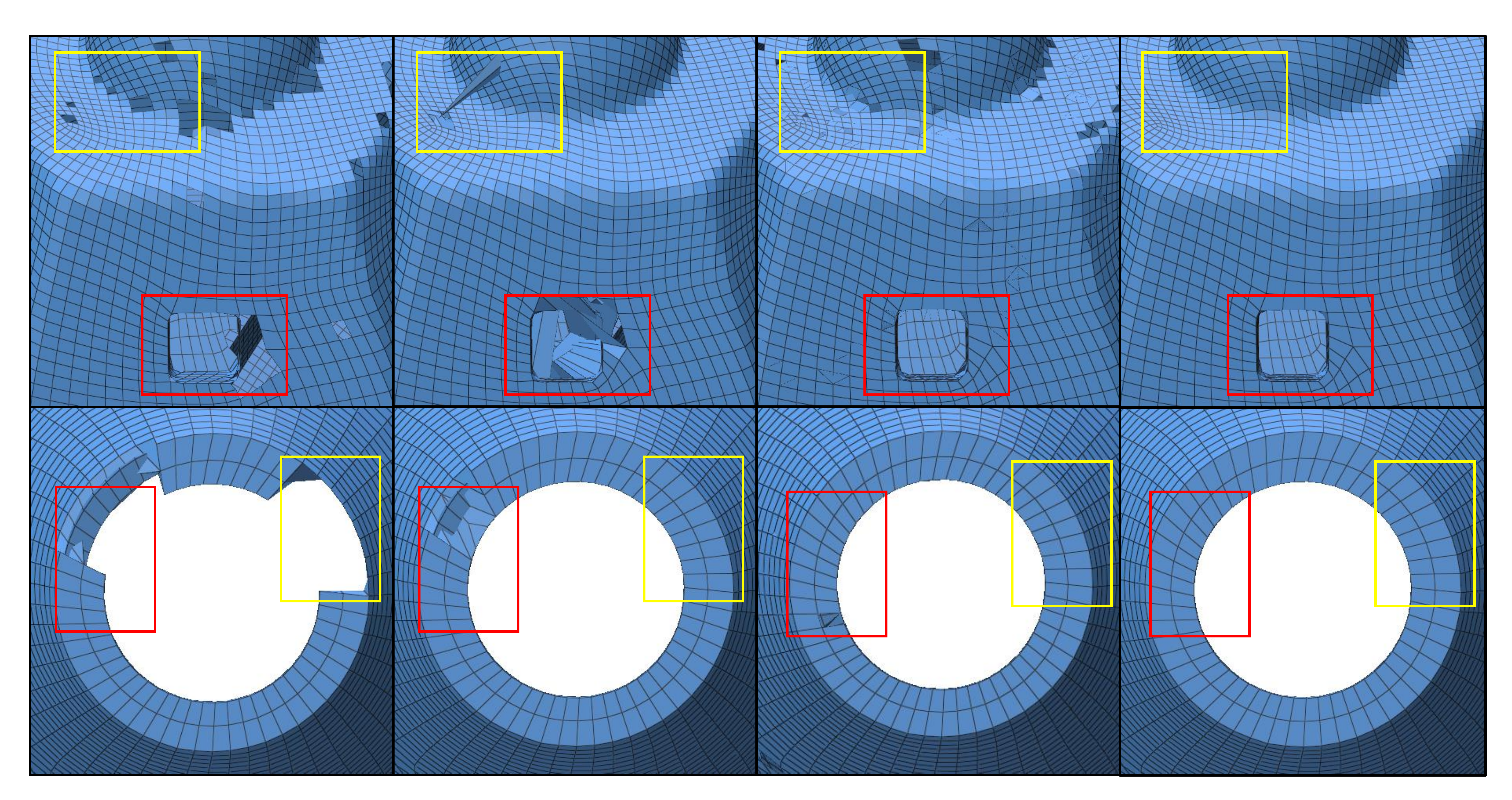}}
\caption{Qualitative demonstration of the ablation study on the designed modules, illustrating their contributions to the quality of the quad mesh.}
\label{fig:ablation}
\vspace{-3mm}
\end{figure}

\subsection{Ablation studies}\label{sec:ablation}
To validate the effectiveness of the core designs within our Point2Quad pipeline for enhancing the quality of output meshes, we conducted ablation studies on the face encoder, face loss, and post-process, and reported the experimental results in Tab.~\ref{tab:ablation}. The results indicate that both components contribute to improving the performance of our model. Removing the face encoder results in a significant decline in mesh quality, with increased angle distortion and Chamfer distance, as well as reduced watertightness. This indicates a higher likelihood of holes forming in the meshes without the face encoder. Furthermore, experimental results reveal that omitting the face loss can lead to \textit{NaN} losses during the later stages of training. The post-process primarily enhances the watertightness and manifold properties of the output meshes, ensuring improved overall quality.

\begin{table}[ht!]
\caption{Ablation studies on pipeline.``w/o FE", ``w/o FL", and ``w/o PP" represent  ``without face encoder", ``without face loss", and ``without post-process". Quantitative comparisons indicate without face encoder, face loss, or post-process, the quality of model's output meshes would deteriorate.}
\footnotesize
\begin{center}
\renewcommand{\tabcolsep}{1.2mm}
\begin{tabular}{c|c|c|c|c}
\hline
Metrics   & w/o FE & w/o FL & w/o PP   & Full  \\
\hline
Scaled Jacobian  & 0.96233 & 0.97072&0.98589 & 0.98613\\
Max-min edge ratio & 16.71773 &2.23763 & 7.91337 & 1.19228\\
Angle distortion  & 9.21026 & 9.09656& 7.54334 & 7.47548\\
Watertightness  & 0.99958 & 0.99911& 0.83587 & 0.99998\\
Manifoldness  & 0.99988 & 0.99988 & 0.99994 & 0.99999\\
Chamfer distance  & 0.04272& 0.02832& 0.01997&0.02064 \\
\hline
\end{tabular}
\label{tab:ablation}
\end{center}
\end{table}

\begin{figure}[htb]
\centering
\includegraphics[width=1.0\linewidth]{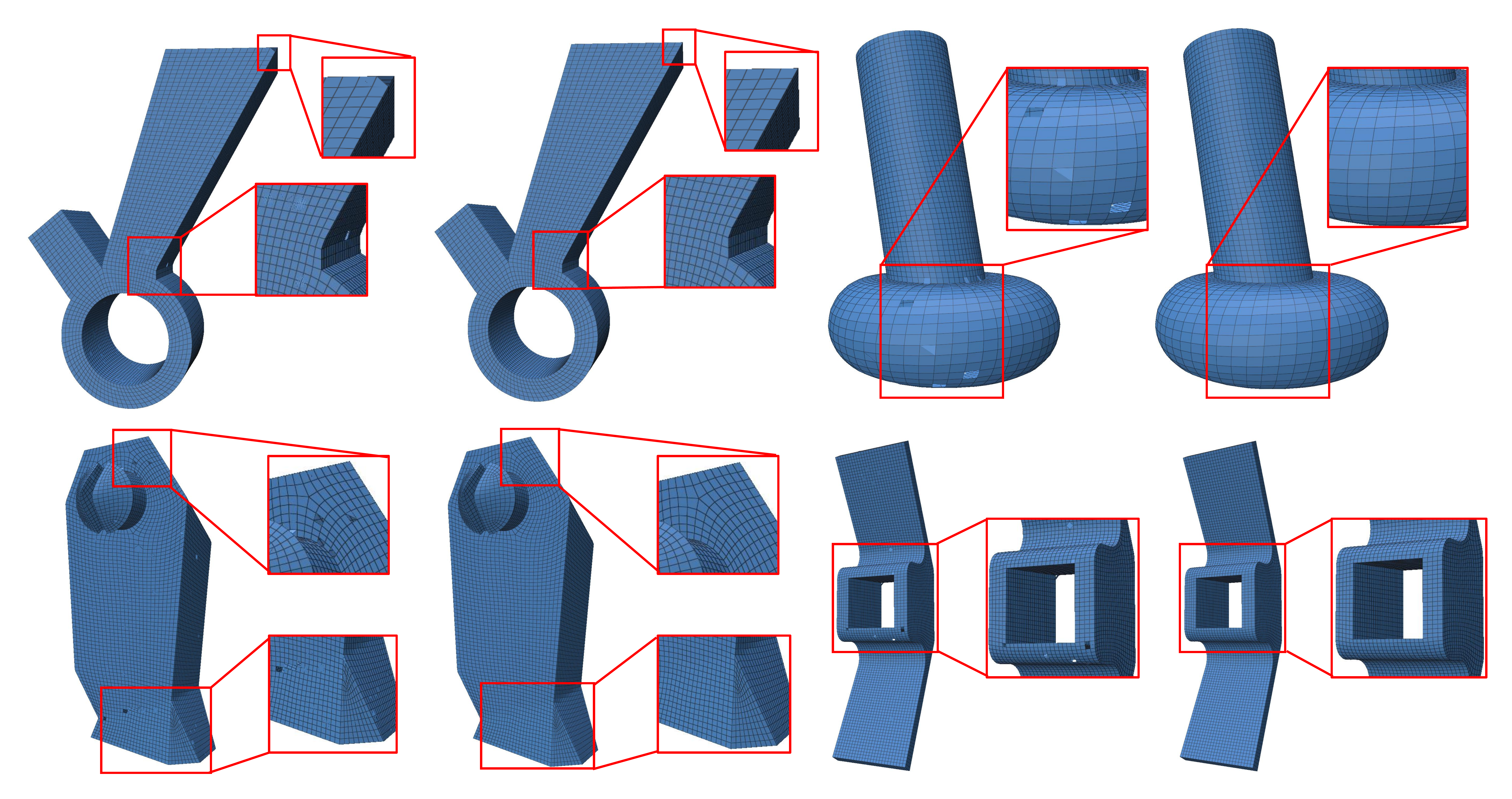}
\vspace{-3mm}
\caption{Quad mesh results: before and after the proposed post-processing.}
\label{fig:bfpp}
\end{figure}

{The ablation study (Fig.~\ref{fig:ablation}) reveals the significant contributions of both the face encoder and the face loss. The former enhances filtering performance, particularly in challenging regions with high curvature or sparse point clouds, while the latter effectively minimizes mesh overlaps and small holes. Our proposed post-processing method effectively addresses non-manifold and non-watertight issues. This is further corroborated by the results presented in Fig.~\ref{fig:bfpp}, which compare the outputs before and after post-processing, showcasing its robust ability to ensure valid mesh topology.}

Furthermore, to assess the impact of each component of $\boldsymbol{\mathcal{F}}_{info}$ on the overall performance of the model, we also conducted ablation studies on the components of $\boldsymbol{\mathcal{F}}_{info}$, including coordinates of points, normal vectors of points, the scaled Jacobian value, and sine values of the four interior angles of a quadrilateral. The experimental results are presented in Tab.~\ref{tab:ablation2}, which indicate that all components of $\boldsymbol{\mathcal{F}}_{info}$ are beneficial for improving the quality of generating quadrilateral meshes. 
Notably, coordinate information significantly impacts the Chamfer distance, maximum-to-minimum edge ratio, and watertightness. Jacobi features strongly affect the scaled Jacobian and Max-min edge ratio, while sine values are pivotal in minimizing angle distortion. The increase in Chamfer distance and reduction in watertightness are directly attributed to the formation of small holes in the meshes.
\begin{table}[ht!]
\caption{Ablation studies on $\boldsymbol{\mathcal{F}}_{info}$.``w/o CO", ``w/o SJ", ``w/o SV", and ``w/o NV" represent  ``without coordinates", ``without scaled Jacobian", ``without sine values", and ``without normal vectors". Quantitative comparisons indicate without CO, SJ, SV, or NV, our model's performance would degrade.}
\footnotesize
\begin{center}
\renewcommand{\tabcolsep}{1.5mm}
\begin{tabular}{@{}c|c|c|c|c|c@{}}
\hline
Metrics   & w/o CO & w/o SJ & w/o SV & w/o NV  & Full  \\
\hline
Scaled Jacobian  & 0.96509 & 0.96493&0.98474 & 0.97307 & 0.98613\\
Max-min edge ratio & 37.22099 &20.9638 &13.67297 & 15.53114 & 1.19228\\
Angle distortion  & 12.50555 &7.46888 &9.55320 & 8.00667 & 7.47548\\
Watertightness  & 0.95065 &0.98676 & 0.96756 & 0.96877 & 0.99998\\
Manifoldness  & 0.99988 & 1.0 & 1.0 & 1.0 &0.99999\\
Chamfer distance  & 0.03723& 0.02095& 0.02197&0.02204 &0.02064 \\
\hline
\end{tabular}
\label{tab:ablation2}
\end{center}
\end{table}

\section{Conclusion}
We have demonstrated \textbf{Point2Quad}, a novel method for generating diverse quad-only meshes from point clouds. Point2Quad regards mesh generation as a face prediction task and utilizes a neural network to select the correct quadrilaterals for each point. Firstly, a point encoder and a face encoder were adopted to extract geometric and local topological features, and then a shared MLP-based classifier was used to filter out the candidate faces. To further improve the watertightness and manifold properties of the generated meshes, we introduced a face loss to guide the learning and performed a simple post-process on the network's output. Our method generated quad meshes with superior quality and shape fitting compared to previous works, as demonstrated by extensive comparative experiments. Ablation studies on the face encoder, face loss, and post-process confirmed the critical contributions of these components to our approach. Furthermore, ablation experiments on the input features of the face encoder demonstrated the importance of all five components in enhancing the quality of the generated quadrilateral meshes.

Despite demonstrating superior reconstruction performance compared to state-of-the-art methods, it cannot guarantee to generate a 100\% watertight and manifold quad mesh, especially for low-quality input point clouds. {Furthermore, our model’s generalization requires improvement, especially when handling sparse or highly non-uniform point clouds, which result in performance degradation. In the future, we will explore quad mesh generation methods with improved generalization capabilities and robust manifoldness guarantees.}

\bibliographystyle{IEEEtran}
\bibliography{ref}

\begin{IEEEbiography}[{\includegraphics[width=0.90in,height=1.20in,clip,keepaspectratio]{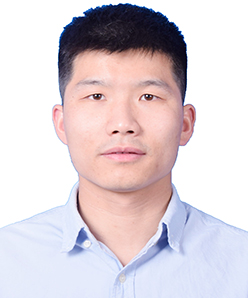}}]{Zezeng Li}
	received a B.S. degree from Beijing University of Technology (BJUT) in 2015 and a Ph.D. degree from Dalian University of Technology (DLUT) in 2024. His research interests include image processing, point cloud processing, and generative models.
\end{IEEEbiography}

\begin{IEEEbiography}[{\includegraphics[width=1.3in,height=1.20in,clip,keepaspectratio]{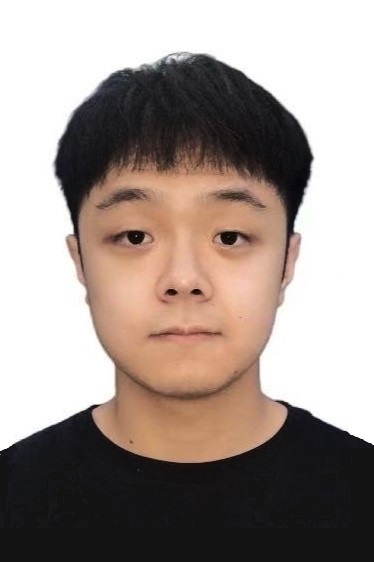}}]{Zhihui Qi}
	received a B.S. degree from Dalian University of Technology (DLUT)  in 2021. He is currently a doctoral student at the School of Software, Dalian University of Technology. His research interests include computer graphics, mesh generation, mesh processing, and computational geometry.
\end{IEEEbiography}

\begin{IEEEbiography}[{\includegraphics[width=0.9in,height=1.4in,clip,keepaspectratio]{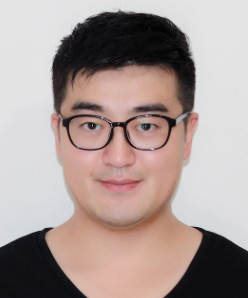}}]{Weimin Wang}
received a B.S. degree from Shanghai Jiao Tong University in 2009 and a Ph.D. degree from Nagoya University in 2017. He is currently an Associate Professor at Dalian University of Technology. Before joining DLUT, he was a researcher at the Artificial Intelligence Research Center of AIST from 2018 to 2021. He researches machine perception, scene understanding and interaction from different scales of the 3D physical world, aiming to enable machines to see the world in super vision.
\end{IEEEbiography}

\begin{IEEEbiography}[{\includegraphics[width=0.9in,height=1.25in,clip,keepaspectratio]{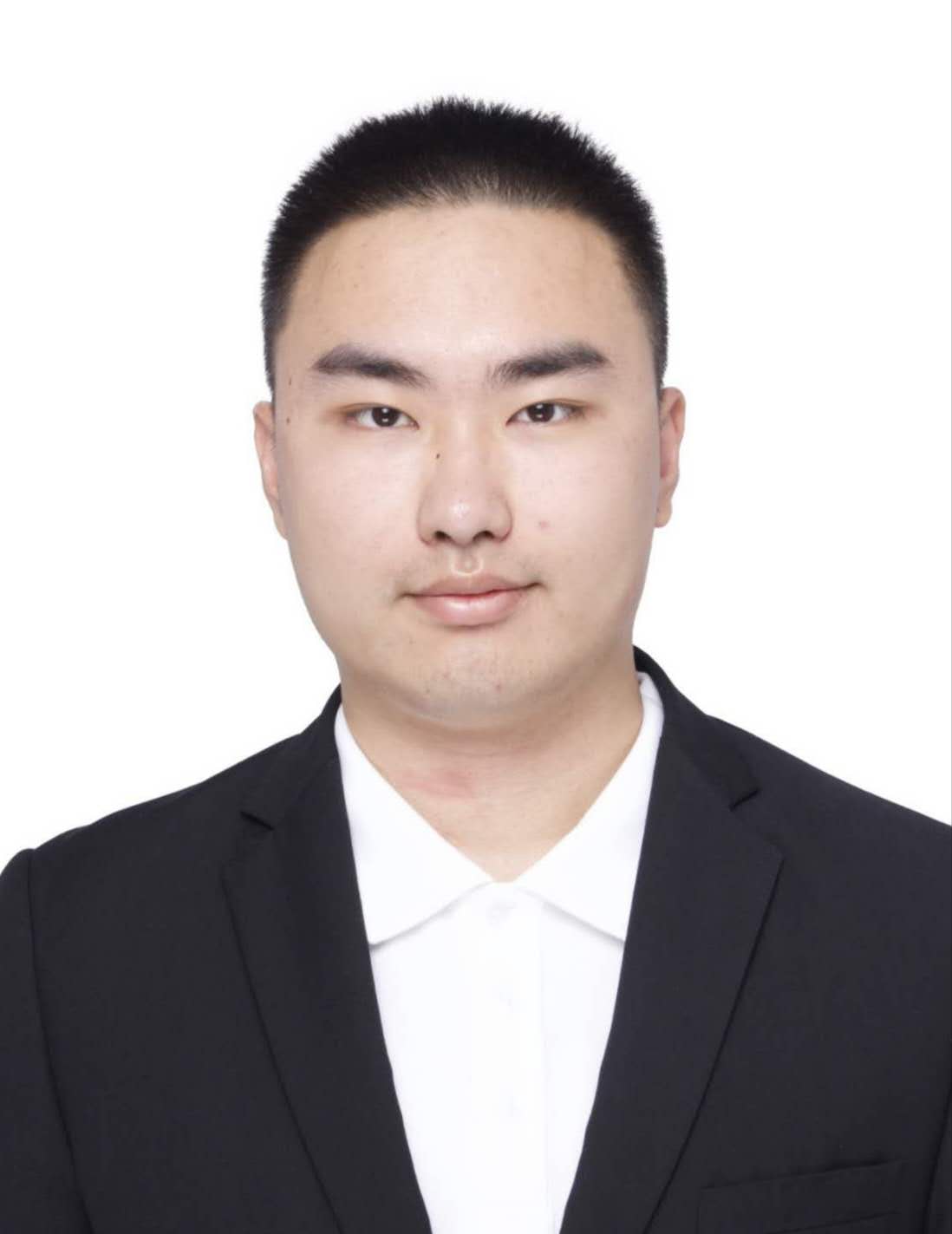}}]{Ziliang Wang} received a B.S. degree from Dalian University of Technology (DLUT) in 2024. He is currently pursuing his graduate studies at Waseda University. His primary research interests include mesh generation and topology optimization.
\end{IEEEbiography}

\begin{IEEEbiography}[{\includegraphics[width=1.2in,height=1.25in,clip,keepaspectratio]{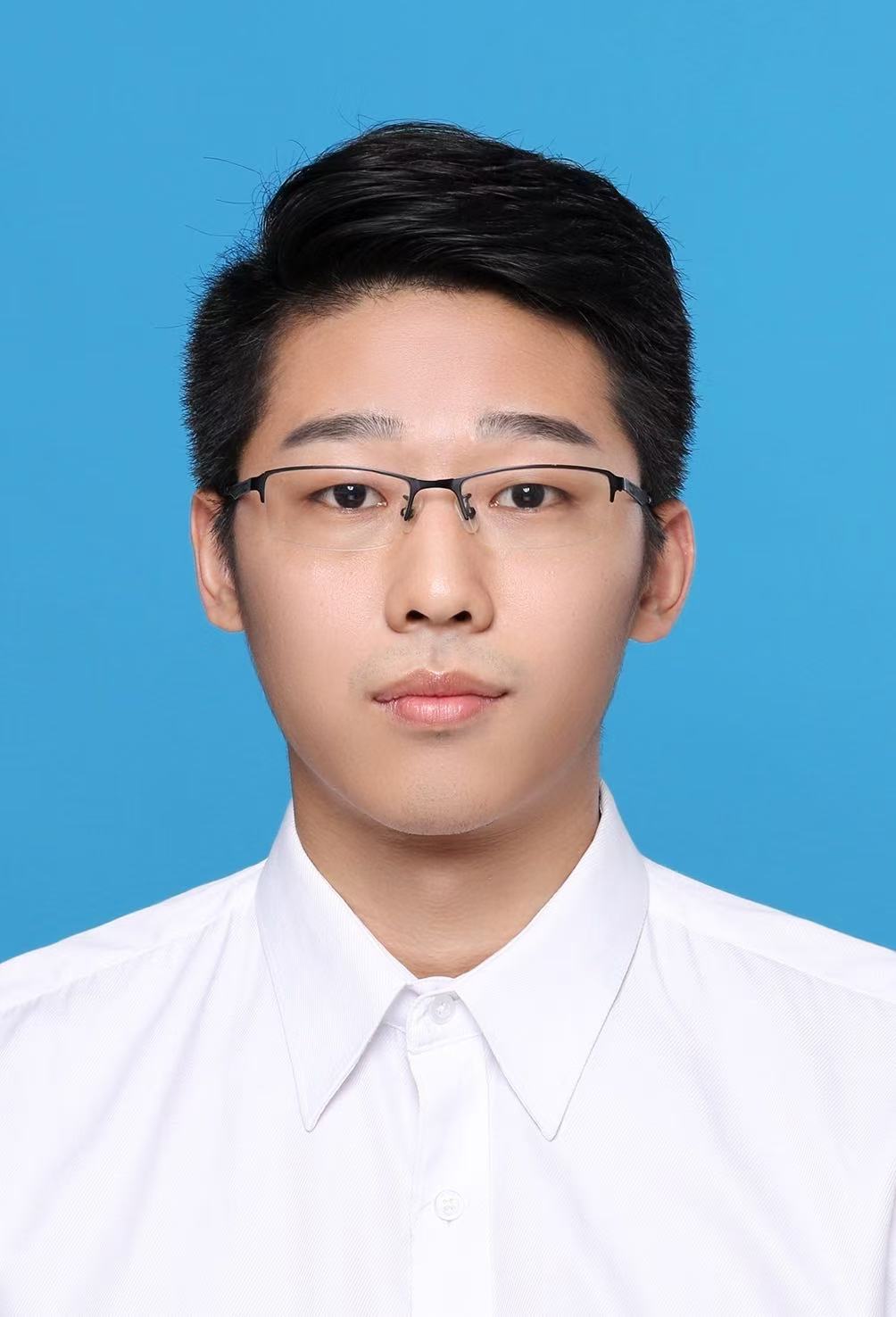}}]{Junyi Duan} obtained his B.S. degree in Computer Science and Technology from Zhengzhou University in 2016, and his Master's degree in Software Engineering from Zhengzhou University in 2019. In 2024, he earned his Ph.D. in Software Engineering from Dalian University of Technology. His main research interests include Computer-Aided Design/Manufacturing, Computer Graphics, and Mesh Generation.
\end{IEEEbiography}

\begin{IEEEbiography}[{\includegraphics[width=1in,height=1.25in,clip,keepaspectratio]{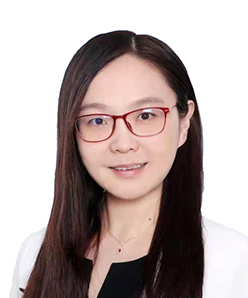}}]{Na Lei}
	received her B.S. degree in 1998 and a Ph.D. degree in 2002 from Jilin University. Currently, she is a professor at Dalian University of Technology. Her research interest is the application of modern differential geometry and algebraic geometry to solve problems in engineering and medical fields. Her work primarily focuses on computational conformal geometry, computer mathematics, and its applications in computer vision and geometric modeling. She is leading a prestigious Chinese National Outstanding Youth Project.
\end{IEEEbiography}

\end{document}